\documentclass[a4paper]{article}

\usepackage{amsmath,amsthm}
\usepackage{pstricks}
\usepackage{pst-node}

\newtheorem{thm}{Theorem}[section]

\newtheorem{lemma}[thm]{Lemma}
\newtheorem{prop}[thm]{Proposition}

\newcommand{\NP}{\ensuremath{\textrm{NP}}}

\newcommand{\node}[1]{\pscirclebox{#1}}
\newcommand{\pre}{\ensuremath{\mathbf{pre}}}
\newcommand{\post}{\ensuremath{\mathbf{post}}}
\newcommand{\prv}{\ensuremath{\mathbf{prv}}}

\author{Omer Gim\'enez \footnote{Llenguatges i Sistemes Inform\`atics,
Universitat Polit\`ecnica de Catalunya,
Barcelona. {\small \tt omer.gimenez@upc.edu }
}}

\title{Solving planning domains with polytree causal graphs is NP-complete}
\date{}

\begin{document}

\maketitle

\abstract{We show that solving planning domains on binary variables
with polytree causal graph is \NP-complete. This is in contrast
to a polynomial-time algorithm of Domshlak and Brafman that solves
these planning domains for polytree causal graphs of bounded
indegree.}

\section{Introduction}

It is well known that the planning problem (namely, the problem of
obtaining a valid sequence of transformations that moves a system from
an initial state to a goal state) is intractable in general
\cite{Chapman87}. However, it is widely believed that many real-life
problems have a particular structure, and that by exploiting this structure
general planners will be able to efficiently handle more meaningful problems.

One of the most fruitful tools researchers have been using to
characterize structure in planning problems is the so called
\emph{causal graph} (\cite{Knoblock}). In short, the causal graph of a problem instance
is a graph that captures the degree of interdependence among the state
variables of the problem.The causal graph has been used both as a tool
for describing tractable subclasses of planning problems (e.g.,
\cite{NASA}, \cite{Carmel-TW}, \cite{Carmel-base}) and as
a key property which algorithms that adress the general planning
problem take into consideration \cite{Helmert}.

In the present work we show that solving planning domains where the
causal graph is a polytree (that is, the underlying undirected graph
is acyclic) is \NP-complete, even if we restrict to domains with
binary variables and unary operators. This result closes the complexity
gap that appears in \cite{Carmel-base}, where it is shown that plan
existence is \NP-complete for planning domains with singly connected
causal graphs, and that plan generation is polynomial for planning
domains with polytree causal graphs of bounded indegree.

Additionally, it is known that solving unary operator planning
problems on binary variables is essentially equivalent to solving
dominance queries for binary-valued CP-nets (see
\cite{Carmel-CP}). Under this reformulation the causal graph becomes the
CP-net, so the present work also shows that dominance testing for
binary-valued polytree CP-nets is NP-complete.

\section{Definitions}\label{sec:def}

In this section we define planning problems and causal graphs according to the
{\sc sas}+ formalism, and we introduce a short-cut notation to describe
unary operators on binary variables.

Let $\mathcal{V}$ be a set of (state) variables. The \emph{domain}
$D_v$ of variable $v$ is the set of values that $v$ can take.  A
\emph{(partial) state} $S$ defined on the set
$\mathcal{C}(S)\subseteq\mathcal{V}$ is a mapping of
the variables $\mathcal{C}(S)$ onto values of their respective domains. When
$\mathcal{C}(S)=\mathcal{V}$ we say that the state $S$ is
\emph{total}. We write $S\subseteq S'$ when
$\mathcal{C}(S)\subseteq\mathcal{C}(S')$ and both assignments coincide
in $\mathcal{C}(S)$, and $S\oplus S'$ to denote the state
defined on $\mathcal{C}(S)\cup\mathcal{C}(S')$ obtained by merging the
assigments of $S$ and $S'$ but giving preference to $S'$ for variables
on $\mathcal{C}(S)\cap\mathcal{C}(S')$.

An \emph{operator} $\alpha$ is a tuple of partial states $(\prv, \pre, \post)$,
where $\prv$ (prevail conditions), $\pre$ (pre-conditions), $\post$
(post-conditions) satisfy $\mathcal{C}(\pre)=\mathcal{C}(\post)$ and
$\mathcal{C}(\prv)\cap\mathcal{C}(\pre)=\emptyset$.
An operator is \emph{unary} when $|\mathcal{C}(\pre)|=1$.
To apply an operator $\alpha$ onto a (total) state $S$ we require
that $\prv\subseteq S$ and $\pre\subseteq S$; when this holds, we
define $\alpha(S)$ as $S\oplus \post$.

A planning domain instance $P$ is a tuple $(\mathcal{V}, \mathcal{O},
I, G)$ where $\mathcal{V}$ is the set of variables, $\mathcal{O}$ is
the set of \emph{operators}, $I$ is the (total) \emph{initial state}
and $G$ is the (possibly partial) \emph{goal state}. A \emph{plan}
$\pi$ for $P$ is a sequence of operators
$\alpha_1,\alpha_2,\cdots,\alpha_t$ such that we are allowed to apply
$\alpha_i$ onto state $S_{i}$ for all $i\leq t$, where
$S_{i}=\alpha_{i-1}(S_{i-1})$ and $S_1=I$, the initial state. An
\emph{action} is a particular occurrence of an operator in a
plan, and a plan is \emph{valid} when $G\subseteq S_{t+1}$.

The \emph{planning problem} is the problem of obtaining a valid plan
$\pi$ for a planning domain instance $P$. We may consider several variations
on the problem, like obtaining optimal valid plans, or
simply deciding whether a plan exists or not.

The \emph{causal graph} of a problem $P=(\mathcal{V},\mathcal{O},I,G)$
is a directed graph that has $\mathcal{V}$ as the set of vertices and
a directed edge from $x$ to $y$ if and only if there is an operator
$\alpha=(\prv, \pre, \post)$ in $\mathcal{O}$ such that
$y\in\mathcal{C}(\post)$ and
$x\in\mathcal{C}(\pre)\cup\mathcal{C}(\prv)$. Hence a directed edge
from $x$ to $y$ means that we may need to take into account the value
of $x$ when considering operators that change $y$.

In the present work we restrict to binary domains (that is,
$\mathcal{D}_v=\{0,1\}$ for all variables $v$) and unary operators
($|\post|=1$). Under these circumstances, and assuming that no
operator has equal pre-condition and post-condition, the pre-condition
of an operator $\alpha$ can be deduced from its post-condition, so we
will simply write $\alpha=\langle \prv, \post \rangle$, or even
$\alpha=\langle \post \rangle$ if $\prv=\emptyset$. In addition, we
write post-conditions using the assignment notation $variable
\leftarrow value$ to emphasize that post-conditions modify the
state. For instance, operators $(\{x=1,y=1\}, \{z=0\}, \{z=1\})$ and
$(\emptyset, \{z=1\}, \{z=0\})$ will be written $\langle \{x=1, y=1\},
z\leftarrow 1\rangle$ and $\langle z\leftarrow 0 \rangle$.

\section{Main result}\label{sec:main}

\begin{figure}[htbp]
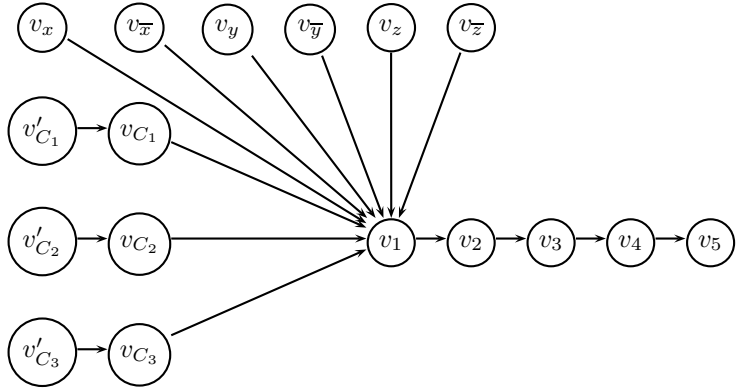


\begin{center}

\begin{psmatrix}[mnode=R, rowsep=0.5cm, colsep=0.4cm]
[name=X] \node{$v_x$} & [name=XN] \node{$v_{\overline{x}}$} & [name=Y] \node{$v_y$} & [name=YN] \node{$v_{\overline{y}}$} &
      [name=Z] \node{$v_z$} & [name=ZN] \node{$v_{\overline{z}}$} & & & & & \\[0pt]
[name=CP1] \node{$v_{C_1}'$} & [name=C1] \node{$v_{C_1}$} & & & & & & & & & \\[0pt]
[name=CP2] \node{$v_{C_2}'$} & [name=C2] \node{$v_{C_2}$} & & &
      [name=V1] \node{$v_1$} & [name=V2] \node{$v_2$} & [name=V3] \node{$v_3$} &
      [name=V4] \node{$v_4$} & [name=V5] \node{$v_5$} \\[0pt]
[name=CP3] \node{$v_{C_3}'$} & [name=C3] \node{$v_{C_3}$} & & & & & & & & &

\psset{arrows=->}

\ncline{X}{V1} \ncline{XN}{V1}
\ncline{Y}{V1} \ncline{YN}{V1}
\ncline{Z}{V1} \ncline{ZN}{V1}
\ncline{V1}{V2}
\ncline{V2}{V3}
\ncline{V3}{V4}
\ncline{V4}{V5}
\ncline{CP1}{C1}
\ncline{CP2}{C2}
\ncline{CP3}{C3}
\ncline{C1}{V1}
\ncline{C2}{V1}
\ncline{C3}{V1}
\end{psmatrix}

\end{center}

\caption{Causal graph of $P_F$ when $F=C_1\wedge C_2\wedge C_3$ on three
variables $x, y, z$.}
\label{fig:reduction}
\end{figure}

We prove NP-hardness by showing a reduction between {\sc 3-CNF-Sat}
and our class of planning domains. As an example of the reduction,
Figure~\ref{fig:reduction} shows the causal graph of the planning
domain $P_F$ that corresponds to a formula $F$ of three variables and
three clauses. (The precise definition of $P_F$ is given in Proposition~\ref{prop:reduction}.)

Let us describe briefly the idea behind the reduction. The planning
domain $P_F$ has two different parts. The first part (state variables
$v_{x}, v_{\overline{x}}, \ldots$, $v_{C_1}, v'_{C_1}, \ldots$, and
$v_1$) depends on the formula $F$, and has the property that a plan
may change the value of $v_1$ from $0$ to $1$ as many times as the
number of clauses of $F$ that a truth assignment can satisfy. However,
this condition on $v_1$ can not be stated as a planning domain
goal. We overcome this difficulty by introducing a gadget (state variables
$v_1,v_2,\ldots,v_t$) that translates it to a regular planning domain
goal.

We describe this last part. Let $P$ be the planning domain
$(\mathcal{V}, \mathcal{O}, I, G)$ where $\mathcal{V}$ is the set of
variables $\{v_1, \ldots, v_{2k-1}\}$, and $\mathcal{O}$ is the set of
$4k-2$ operators $\{\alpha_1,\ldots, \alpha_{2k-1},\beta_1,\ldots,
\beta_{2k-1}\}$. Operators $\alpha_1$ and $\beta_1$ are defined as
$\langle v_1\leftarrow 0 \rangle$ and $\langle v_1\leftarrow 1
\rangle$; for $i>1$, operators $\alpha_i$ and $\beta_i$ are
respectively $\langle \{v_{i-1}=0\}, v_{i}\leftarrow 0 \rangle$ and
$\langle \{v_{i-1}=1\}, v_{i}\leftarrow 1\rangle$. All variables are
$0$ in the initial state $I$, and the goal state $G$ is $v_i=0$ when
$i$ is even, $v_i=1$ when $i$ is odd.

\begin{lemma}\label{lemma:P}
Any valid plan for the planning domain $P$ changes at least $k$
times the variable $v_1$ from $0$ to $1$. There is a valid plan that
achieves this minimum.
\end{lemma}

\begin{proof}
Let $A_i$ be the sequence of actions $\alpha_1,\ldots, \alpha_i$, and let
$B_i$ be the sequence of actions $\beta_1, \ldots, \beta_i$. It is
easy to check that the plan $B_{2k-1}$, $A_{2k-2}$,$B_{2k-3}$~,~$\ldots$,
$B_3$,$A_2$,$B_1$ is valid: after finishing a sequence of actions $A_{i}$
or $B_{i}$, the variable $v_i$ is in its goal state ($0$
if $i$ is even, $1$ if $i$ is odd). Subsequent actions in the plan
do not modify $v_i$, so the variable remains in its goal state until the
end. The action $\beta_1$ appears $k$ times in the plan, thus $v_1$
changes $k$ times from state $0$ to $1$.

We proceed to show that $k$ is the minimum. Consider a valid plan
$\pi$, and let $\lambda_i$ be the number of actions $\alpha_i$ and
$\beta_i$ that appear in $\pi$. (That is, $\lambda_i$ is the number of
times that variable $v_i$ changes value, either from $0$ to $1$ or
from $1$ to $0$. Note that the number of actions $\beta_i$ has to be
either equal or exactly one more than the number of actions
$\alpha_i$.) We will show that $\lambda_{i-1}>\lambda_{i}$. Since
$\lambda_{2k-1}$ has to be at least one, $\lambda_{i-1}>\lambda_{i}$
implies that $\lambda_1\geq 2k-1$. In consequence, there are at least
$k$ actions $\beta_i$ in plan $\pi$, finishing the proof.

We show that $\lambda_{i-1}>\lambda_{i}$ for valid plans. To begin
with, let $\pi$ be any plan (not necessarily a valid one) and consider
only the subsequence made out of actions $\alpha_i$ and $\beta_i$ in $\pi$.
It starts with $\beta_i$ (since the initial state is $v_i=0$), and the
same action can not appear twice consecutively in the sequence. Thus
this sequence alternates $\beta_i$ and $\alpha_i$.
 Moreover, since $\beta_i$ (for $i>1$) has $v_{i-1}=1$ as
prevail condition, and $\alpha_i$ has $v_{i-1}=0$, there must be at
least one action $\alpha_{i-1}$ in the plan $\pi$ betweeen any two
actions $\beta_i$ and $\alpha_i$. For the same reason we must have at
least one action $\beta_{i-1}$ between any two actions $\alpha_i$ and
$\beta_i$, and an action $\beta_{i-1}$ before the first action
$\beta_i$. This shows that, in any plan $\pi$, not necessarily valid,
we have $\lambda_{i-1}\geq\lambda_{i}$. If, in addition, $\pi$ is
valid, we require an extra action: when $v_i$ changes state for the last
time and attains its goal state, we have that $v_{i-1}=v_i$, so
$v_{i-1}$ is not in its goal state by parity. Hence a valid plan must have an
extra action $\alpha_{i-1}$ or $\beta_{i-1}$ after all occurrences of
$\alpha_i$ and $\beta_i$. Thus $\lambda_{i-1}>\lambda_i$ for valid
plans.
\end{proof}

\begin{prop}\label{prop:reduction}
3-{\sc Sat} reduces to plan existance for planning domains on binary
variables with a polytree causal graph.
\end{prop}

\begin{proof}
Let $F$ be a CNF formula with $k$ clauses and $n$ variables. We produce
a planning domain $P_F$ on $2n+4k-1$ state variables and
$2n+14k-3$ operators.  The planning problem has two state variables $v_x$
and $v_{\overline{x}}$ for every variable $x$ in $F$, two state variables
$v_C$ and $v_C'$ for every clause $C$ in $F$, and $2k-1$ additional
variables $v_1,\ldots,v_{2k-1}$. All variables are $0$ in the initial state. The
(partial) goal state is $v_i=0$ when $i$ is even, $v_i=1$ when $i$ is odd, like
in problem $P$ of Lemma~\ref{lemma:P}. The operators are:

\begin{enumerate}
\item[(1.)] Operators $\langle v_x\leftarrow 1 \rangle$ and
$\langle v_{\overline{x}}\leftarrow 1\rangle$ for every variable $x$ of $F$.
\item[(2.)] Operators $\langle v_C'\leftarrow 1\rangle$,
$\langle \{v_C'=0\}, v_C\leftarrow 1\rangle$ and
$\langle \{v_C'=1\}, v_C\leftarrow 0\rangle$ for every clause $C$ of $F$.
\item[(3.)] There are 7 operators for every clause $C$, one for each
of the 7 different partial assignments that satisfy $C$. The
post-condition of these operators is $v_1\leftarrow 1$, and they all
have 7 prevail conditions: the condition $v_C=1$, and six conditions
to ensure that the values of the state variables $v_x$ and
$v_{\overline{x}}$ associated to a variable $x$ that appears in $C$
are in agreement with the partial assignment.

For example, the operator related to the clause $C=x\vee \overline{y} \vee \overline{z}$
and the satisfying partial assigment $\{x=0, y=0, z=1\}$ is
$$
\langle \{v_x=0, v_{\overline{x}}=1, v_y=0, v_{\overline{y}}=1, v_z=1, v_{\overline{z}}=0, v_C=1\},
 v_1\leftarrow 1 \rangle
$$ 
\item[(4.)] An operator $\langle \{v_C=0\mid\forall C\}, v_1 \leftarrow 0 \rangle$.
\item[(5.)] Operators $\alpha_i=\langle \{v_{i-1}=0\}, v_{i}\leftarrow 0\rangle$
and $\beta_i=\langle \{v_{i-1}=1\}, v_{i}\leftarrow 1\rangle$ for $2\leq i \leq 2k-1$. (That is,
the same operators that in problem $P$ but for $\alpha_1$ and $\beta_1$.) 
\end{enumerate}

We note some simple facts about problem $P_F$. For any variable $x$, state
variables $v_x$ and $v_{\overline{x}}$ in $P_F$ start at $0$, and by using
the actions in (1.), they can change into $1$, but they
can not go back to $0$.  In particular, a plan $\pi$ can not
reach both partial states $\{v_x=1, v_{\overline{x}}=0\}$ and
$\{v_x=0, v_{\overline{x}}=1\}$ during the course of its execution.

Similarly, if $C$ is a clause of $F$, the state variable $v_C$ can change from $0$
to $1$ and, by first changing $v_C'$ into $1$, $v_C$ can go back to $0$.
No further changes are possible, since no action brings back $v_C'$ to $0$.

Now we interpret actions in (3.) and (4.), which are the only actions
that affect $v_1$. To change $v_1$ from $0$ to $1$ we need to apply
one of the actions of (3.), thus we require $v_C=1$ for a clause
$C$. But the only way to bring back $v_1$ to $0$ is applying the
action (4.), so that $v_C=0$. We deduce that every time that $v_1$
changes its value from $0$ to $1$ and then back to $0$ in the plan
$\pi$, at least one of the $k$ state variables $v_C$ is \emph{used
up}, in the sense that $v_C$ has been brought from $0$ to $1$ and then
back to $0$, and cannot be used again for the same purpose.

We show that $F$ is in {\sc 3-Sat} if and only if there is a valid
plan for problem $P_F$. Let $\sigma$ be a truth assignment that
satisfies $F$. By Lemma~\ref{lemma:P} we can extend a plan $\pi'$ that
switches variable $v_1$ from $0$ to $1$ at least $k$ times to a plan
$\pi$ that sets all variables $v_i$ to their goal values. The plan
$\pi'$ starts by setting the all state variables $v_x$ and $v_{\overline{x}}$
in correspondence with the truth assignment $\sigma$ using the actions
of (1.). Then, for every of the $k$ state variables $v_C$, we set
$v_C=1$, we apply the action of (3.) that corresponds to $\sigma$
restricted to the variables of clause $C$, and we move back $v_C$ to
$0$ so that we can apply the action (4.). By repeating this process for
all clauses $C$ of $F$ we are switching the state variable $v_1$
exactly $k$ times from $0$ to $1$.

We show the converse, namely, that the existence of a valid plan $\pi$
in $P_F$ implies that $F$ is satisfiable. By Lemma~\ref{lemma:P} the
state variable $v_1$ has to change from $0$ to $1$ at least $k$
times. This implies that $k$ actions of (3.), all of them
corresponding to different clauses, have been used to move $v_1$ from
$0$ to $1$. Hence we can define a satisfying assigment $\sigma$ by setting
$\sigma(x)=1$ if the partial states $\{v_x=1, v_{\overline{x}}=0\}$ appears during
the execution of $\pi$, and $\sigma(x)=0$ otherwise.
\end{proof}

\begin{thm}\label{thm:reduction}
Plan existence for planning problems with a polytree causal graph is \NP-complete.
\end{thm}

\begin{proof}
Due to Proposition~\ref{prop:reduction} we only need to show that the
problem is \NP. But in \cite{Carmel-base} it is shown that this holds
in the more general setting of planning problems with causal graphs
where each component is singly connected. Their proof uses the
non-trivial result that solvable planning problems on binary variables
with a singly connected causal graph have plans of polynomial
length. (This is not true for non-binary variables, or unrestricted
causal graphs.)

\end{proof}

\section{Concluding remarks}\label{sec:background}

The given reduction constructs a planning domain with a partial goal
state. If desired, we can make the goal state total by adding the
restrictions $v_x=1, v_{\overline{x}}=1$ and $v_C=1, v_C'=0$ for all
variables $x$ and clauses $C$. On the other hand, the author has found
no way to avoid an operator like the one in (4.), with an unbounded
number of prevail conditions. Hence we should not rule out the
existence of a polynomial-time algorithm that solves polytree planning
domains where all operators have prevail conditions bounded by a constant.

Planning domains with unary operators on binary variables can be
formulated as dominance queries in binary CP-nets, and the translation
is polynomial-size preserving provided two technical conditions (see
\cite{Carmel-CP} for the details): we must allow partially specified
CPTs in the CP-net, and no two operators with opposing post-conditions
may share the same prevail conditions in the planning domain. Clearly
the given reduction satisfies this last condition (actions $\alpha_1$
and $\beta_1$ in problem P do not satisfy the last requirement, but
they are not present in problem $P_F$) hence dominance testing in
binary polytree CP-nets (with partially specified CPTs) is
NP-complete.

Brafman and Domshlak show in \cite{Carmel-base} that, if we restrict to
binary variables and unary operators, we can generate valid plans in
time roughly $O(|\mathcal{V}|^{2\kappa})$ for planning domains where
the causal graph is a polytree with indegree bounded by $\kappa$.
The same authors show in \cite{Carmel-TW} how to solve 
in time roughly $O(|V|^{\omega\delta})$ planning domains with local depth
$\delta$ and causal graphs of tree-width $\omega$.
The planning domains of our reduction have $\kappa=2n+k$, $\omega=1$
and $\delta=2k-1$, where $n$ and $k$ stand for the number of variables
and clauses of the 3-CNF formula. This vindicates the fact that the
algorithms are exponential in, respectively, $\kappa$ and
$\delta$, so that, unless $P=NP$, we cannot hope to improve them in a
significant way for polytree planning domains.

\subsection*{Acknowledgements}

The author is in debt to Carmel Domshlak and Anders Jonsson for useful
suggestions and careful reading of this paper.

\end{document}